\documentclass[twoside,11pt,hyperref]{article}

\usepackage{jmlr2e}

\usepackage{amsmath}
\usepackage{amsfonts}
\usepackage{hyperref}[6.83]
\hypersetup{colorlinks,
			linkcolor=[rgb]{.61,0,0.3},
			citecolor=[rgb]{.14,.47,.14}}

\usepackage[capitalize]{cleveref}[0.19]
\usepackage{booktabs}
\usepackage{program}
\NumberProgramstrue

\newtheorem{cor}{Corollary}
\newtheorem{problem}{Problem}
\newtheorem{rmrk}{Remark}

\crefname{table}{Table}{Tables}
\crefname{definition}{Definition}{Definitions}
\crefname{assumption}{Assumption}{Assumptions}
\crefname{theorem}{Theorem}{Theorems}
\crefname{rmrk}{Remark}{Remarks}
\crefname{lemma}{Lemma}{Lemmas}
\crefname{cor}{Corollary}{Corollaries}
\crefname{proposition}{Proposition}{Propositions}
\crefname{section}{Section}{Sections}
\crefname{subsection}{Subsection}{Subsections}
\crefname{example}{Example}{Examples}
\crefname{problem}{Problem}{Problems}

\crefformat{equation}{\textup{#2(#1)#3}}
\crefrangeformat{equation}{\textup{#3(#1)#4--#5(#2)#6}}
\crefmultiformat{equation}{\textup{#2(#1)#3}}{ and \textup{#2(#1)#3}}
{, \textup{#2(#1)#3}}{, and \textup{#2(#1)#3}}
\crefrangemultiformat{equation}{\textup{#3(#1)#4--#5(#2)#6}}%
{ and \textup{#3(#1)#4--#5(#2)#6}}{, \textup{#3(#1)#4--#5(#2)#6}}{, and \textup{#3(#1)#4--#5(#2)#6}}

\Crefformat{equation}{#2Equation~\textup{(#1)}#3}
\Crefrangeformat{equation}{Equations~\textup{#3(#1)#4--#5(#2)#6}}
\Crefmultiformat{equation}{Equations~\textup{#2(#1)#3}}{ and \textup{#2(#1)#3}}
{, \textup{#2(#1)#3}}{, and \textup{#2(#1)#3}}
\Crefrangemultiformat{equation}{Equations~\textup{#3(#1)#4--#5(#2)#6}}%
{ and \textup{#3(#1)#4--#5(#2)#6}}{, \textup{#3(#1)#4--#5(#2)#6}}{, and \textup{#3(#1)#4--#5(#2)#6}}

\crefdefaultlabelformat{#2\textup{#1}#3}

\newcommand{\norm}[1]{\left\Vert #1 \right\Vert}

\newcommand{\Prb}[1]{\mathbb{P}\left[ #1 \right]}
\newcommand{\E}[1]{\mathbb{E}\left[ #1 \right]}
\newcommand{\tr}[1]{\mathbf{tr}\left[  #1 \right]}
\newcommand{\bigO}[1]{\mathcal{O}\left[ #1 \right]}

\newcommand{\cond}[2]{\mathbb{E}\left[\left. #1 \right\vert #2 \right]}

\newcommand{\V}[1]{\mathbb{V}\left[ #1 \right]}

\newcommand{\ra}{\renewcommand{\arraystretch}{1.2}}

\ShortHeadings{Characterizing and Overcoming Stalling}{Patel}
\firstpageno{1}

\begin{document}

\title{On SGD's Failure in Practice: Characterizing and Overcoming Stalling}

\author{\name Vivak Patel \email vp314@uchicago.edu \\
       \addr Department of Statistics\\
       University of Chicago\\
       Chicago, IL 60637, USA}

\editor{} 

\maketitle

\begin{abstract}
Stochastic Gradient Descent (SGD) is widely used in machine learning problems to efficiently perform empirical risk minimization, yet, in practice, SGD is known to stall before reaching the actual minimizer of the empirical risk. SGD stalling has often been attributed to its sensitivity to the conditioning of the problem; however, as we demonstrate, SGD will stall even when applied to a simple linear regression problem with unity condition number for standard learning rates. Thus, in this work, we numerically demonstrate and mathematically argue that stalling is a crippling and generic limitation of SGD and its variants in practice. Once we have established the problem of stalling, we generalize an existing framework for hedging against its effects, which (1) deters SGD and its variants from stalling, (2) still provides convergence guarantees, and (3) makes SGD and its variants more practical methods for minimization.
\end{abstract}


\section{Introduction}
Stochastic Gradient Descent (SGD) is a popular, simple technique for attempting empirical risk minimization, which is defined as 
\begin{equation} \label{problem: Empirical Risk Minimization}
\arg\min_{\beta} \frac{1}{N} \sum_{i=1}^N l(Z_i,\beta) =: \arg\min_{\beta} R_N(\beta)
\end{equation}
where $l: \mathbb{R}^p \times \mathbb{R}^d \rightarrow \mathbb{R}$ is a continuously differentiable, loss function, $Z_1,\ldots,Z_N \in \mathbb{R}^p$ are examples, and $\beta \in \mathbb{R}^d$ is some unknown parameter which must be learned. SGD tries to solve \cref{problem: Empirical Risk Minimization} by producing a sequence of iterates $\lbrace \theta_k : k +1 \in \mathbb{N} \rbrace \subset \mathbb{R}^d$ determined by
\begin{equation} \label{method: SGD}
\theta_{k+1} = \theta_k - \alpha_k \dot{l}(Z_{S_{k+1}},\theta_k)
\end{equation}
where $\theta_0$ is arbitrary, $\dot{l}$ is the gradient of $l$ with respect to $\beta$, $S_k$ is an independently drawn random variable taking values in $1,\ldots,N$, and $\alpha_k > 0$ is the learning rate. When SGD's learning rate satisfies the condition of \cite{robbins1951},
\begin{equation} \label{condition: Robbins-Monro}
\sum_{k=0}^\infty \alpha_k = \infty \quad\text{and}\quad \sum_{k=0}^\infty \alpha_k^2 < \infty
\end{equation}
SGD converges in theory \citep[Proposition 4.3]{bertsekas2010}. Moreover, if SGD's learning rate is $\bigO{(k+1)^{-1}}$ then 
\begin{equation} \label{equation: Objective Expected Convergence Rate} 
\E{ R_N(\theta_k)} - R_N(\beta^*) = \bigO{ \frac{\rho\kappa^2}{k} }
\end{equation}
where $\beta^*$ is the minimizer of the empirical risk, $\rho$ is a problem dependent parameter, and  $\kappa$ is the condition number of $\nabla^2 R_N(\beta^*)$ \citep{murata1998}.

SGD's severe sensitivity to the conditioning has often been cited as the reason for its relatively slow (or lack of) convergence in practice, and has inspired a number of adaptive learning rate variants \citep{duchi2011,tieleman2012,zeiler2012,konevcny2013,schaul2013,kingma2014} and second order variants \citep{amari2000,schraudolph2007,bordes2009,sohl2013,byrd2016,patel2016} to overcome this short-coming. However, these SGD variants fail to recognize and address a more general, deeper problem known to practitioners: stalling -- a phenomenon in which a theoretically convergent method stops making meaningful progress to the solution after a period of time. 

In this work, we endeavor to characterize the phenomenon of stalling, highlight its mechanism, demonstrate its genericness, and provide a strategy  to hedge against its effects. In particular,
\begin{enumerate}
\item in \cref{section: Motivating Problem}, we state an ideal risk minimization problem and give theoretical and experimental characterizations of SGD on this problem, which we use to demonstrate the phenomenon of stalling. We then describe the mechanisms of stalling.
\item in \cref{section: Restart Framework}, we generalize a strategy of restarting to hedge against stalling, and state an algorithm for restarting SGD. Our strategy extends the work of \cite{loshchilov2016}, which is the first to numerically explore restarted SGD in the context of deep neural networks.  We further this line of work by stating a restart strategy which can be applied to arbitrary stochastic gradient methods with random restart points, and by proving sufficient conditions for convergence of restarted SGD.  
\item in \cref{section: Experiments}, we compare the standard and our restarted variants of SGD, AdaGrad \citep{duchi2011}, and kSGD \citep{patel2016}, for fitting a neural network to classify electron neutrinos and muon neutrinos from data generated by a Fermi Lab experiment \citep{roe2005}. For comparison, we also fit the neural network using BFGS \citep[Chapter 6]{nocedal2006}.
\end{enumerate}
In \cref{section: Conclusion}, we summarize this work and mention some future directions. In the appendices, we prove the results stated in the preceding sections.

\section{Stalling} \label{section: Motivating Problem}
In \cref{subsection: Problem Definition}, we define a simple linear regression problem which has a condition number of one. In \cref{subsection: Convergence Theory Stalling Practice}, we demonstrate numerically and mathematically that SGD will stall for such a problem despite theoretical convergence guarantees. In \cref{subsection: general problem}, we formulate a strongly convex, Lipschitz continuous risk minimization problem which will serve as the theoretical framework for the remainder of this work. In \cref{subsection: mechanisms}, using the more general framework, we discuss the mechanism behind stalling.

\subsection{Ideal Problem Definition} \label{subsection: Problem Definition}
Let $Q \in \mathbb{R}^{d \times d}$ be an orthonormal matrix. Let $X,X_1,X_2,\ldots \in \mathbb{R}^d$ be independent random vectors which are drawn from the columns of $Q$ such that each column has an equal probability of being drawn. Now, let $\epsilon, \epsilon_1,\epsilon_2,\ldots \in \mathbb{R}$ be independent, identically distributed, symmetric, bounded random variables with mean zero. Let $\beta^* \in \mathbb{R}^d$ be an arbitrary vector, and define 
\begin{equation}
Y = X'\beta^* + \epsilon \quad\text{and}\quad Y_i = X_i'\beta^* + \epsilon_i \quad\forall i \in \mathbb{N}
\end{equation}

\begin{problem}[Ideal Problem] \label{problem: Ideal Problem}
Given observations $(Y_1,X_1),(Y_2,X_2),\ldots$, compute $\beta^*$. 
\end{problem}

There are several reasons which make \cref{problem: Ideal Problem} the ideal problem. First, all of \cref{problem: Ideal Problem}'s random variables are bounded, which prevents extremal realizations of the random variables from derailing the SGD estimates. Second, \cref{problem: Ideal Problem} includes \cref{problem: Empirical Risk Minimization} as a special case for techniques which are online or rely only on subsampling. Third, \cref{problem: Ideal Problem} defines a simple strongly convex, differentiable, quadratic problem for the commonly used squared error loss,
\begin{equation} \label{eqn: Expected Risk for Linear Regression}
R(\beta) = \E{ (Y - X'\beta)^2} = \V{\epsilon} + \frac{1}{d}\norm{\beta - \beta^*}_2^2
\end{equation}
Finally, as evidenced by \cref{eqn: Expected Risk for Linear Regression}, \cref{problem: Ideal Problem} has a condition number of $1$. Thus, \cref{problem: Ideal Problem} ought to be the ideal problem for SGD to solve. Moreover, because \cref{problem: Ideal Problem} does not introduce any non-linearity issues or curvature difficulties, any adaptive learning rate or second-order SGD variant should have no additional benefit in comparison to SGD on this simple problem.

\subsection{Converging in Theory, Stalling in Practice} \label{subsection: Convergence Theory Stalling Practice}

If SGD is applied to solve \cref{problem: Ideal Problem}, then $\lbrace \theta_k : k +1 \in \mathbb{N} \rbrace$ are, using \cref{method: SGD}, 
\begin{equation}
\theta_{k+1} = \theta_k + \alpha_k X_{k+1} \left(Y_{k+1} - X_{k+1}'\theta_k \right)
\end{equation}
where $\theta_0 \in \mathbb{R}^d$ is arbitrary. Then,
\begin{theorem} \label{theorem: SGD Convergence for Ideal Problem}
For any non-negative sequence $\lbrace \alpha_k \rbrace$  
$$
\begin{aligned}
\E{R(\theta_k)} - R(\beta^*) &= \left[ R(\theta_0) - R(\beta^*) \right] \prod_{j=0}^k \left(1 + \frac{\alpha_j^2 - 2\alpha_j}{d} \right) \\
& + \frac{\V{\epsilon}}{d}\left[ \alpha_k^2 + \sum_{j=0}^{k-1} \alpha_j^2 \prod_{l=j+1}^k \left(1 + \frac{\alpha_l^2 - 2\alpha_l}{d} \right) \right]  \end{aligned} $$
Thus, $\theta_k \to \beta^*$ in probability if $\limsup A_k = 0$, where  
$$ A_k :=  \alpha_k^2 + \sum_{j=0}^{k-1} \alpha_j^2 \prod_{l=j+1}^k \left(1 + \frac{\alpha_l^2 - 2\alpha_l}{d} \right)$$
\end{theorem}
In particular, if SGD's learning rate satisfies \cref{condition: Robbins-Monro}, then $\theta_k \to \beta^*$ in probability. For example, if SGD's learning rate is of the form 
\begin{equation} \label{equation: Example Learning Rates}
\alpha_{k} = (k+1)^{-p}
\end{equation}
where $p \in (0.5,1]$, which are typical choices in practice, then $\theta_k \to \beta^*$ in probability. 

\begin{figure}[ht]
\centering
\includegraphics[scale=0.7]{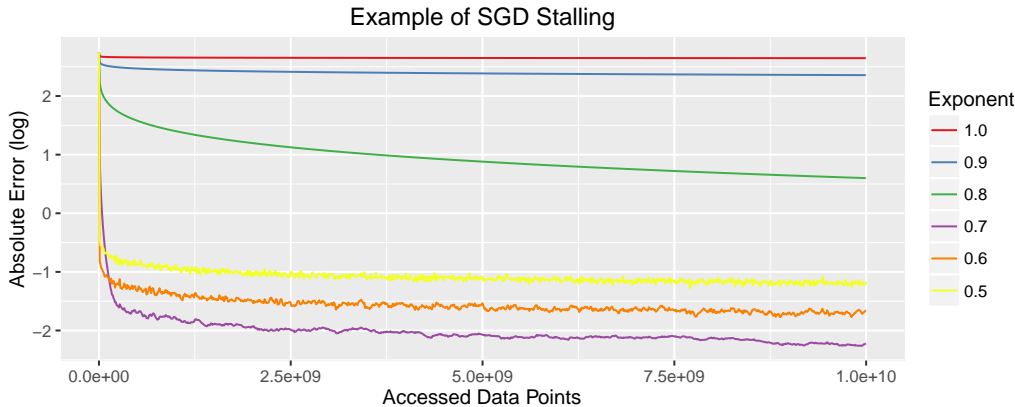}
\caption{The base-10 logarithm of the absolute error between the estimator generated by SGD with $\theta_0 = 0 \in \mathbb{R}^{100}$ and the true parameter for \cref{problem: Ideal Problem} with $\epsilon \sim Unif(-5,5)$. The learning rates are selected to be of the form in \cref{equation: Example Learning Rates} with $p = 0.5, 0.6, 0.7, 0.8, 0.9, 1.0$. Each SGD run sees exactly the same ten billion observations in the same order. Despite the number of observations and excellent conditioning, the logarithm of the absolute error remains larger than $-2.5$ at best.}
\label{figure:SGD Stalling}
\end{figure}

Unfortunately, SGD's convergence in probability does not translate into convergence in practice: on the experiment described in \cref{figure:SGD Stalling}, SGD fails to continue making meaningful progress to the optimal point after ten billion observations despite the near one condition number. As this experiment demonstrates, there is a disconnect between the theoretical behavior of SGD and its practical performance. In order to reconcile the theory and practice, we must give a quantitative result about how quickly $\theta_k$ converges to $\beta^*$ in probability. One such example is given in \cref{corollary: SGD Convergence Rate Lower Bound}.

\begin{cor} \label{corollary: SGD Convergence Rate Lower Bound}
For $\delta > 0$, 
$$\Prb{\norm{\theta_k - \beta^*}_2 \leq \delta} \geq 1 - \frac{\norm{\theta_0 - \beta^*}_2^2 \prod_{j=0}^k \left(1 + \frac{\alpha_j^2 - 2\alpha_j}{d} \right) + \V{\epsilon}A_k}{\delta^2}$$
\end{cor}

If \cref{corollary: SGD Convergence Rate Lower Bound} is applied to \cref{problem: Ideal Problem}, then a lower bound for the probability that $\norm{\theta_k - \beta^*}_2$ is smaller than $10^{-1}$ is tabulated in \cref{table:linear problem lower probability bounds} for the problem in \cref{figure:SGD Stalling}. Even for the generous error bound of $10^{-1}$, \cref{table:linear problem lower probability bounds} suggests that SGD performs rather poorly in practice. If the error bound is decreased to $10^{-3}$, then, even with ten billion observations, the lower probability bounds would all remain $0.0$. Thus, \cref{table:linear problem lower probability bounds} paints a rather pessimistic picture of SGD's performance in practice. 

\begin{table}[ht]
\centering
\ra
\begin{tabular}{@{}c c rrrrrr@{}}\toprule
\multicolumn{8}{c}{SGD Convergence Probability Lower Bound, $\delta = 10^{-1}$, $d=100$} \\ \midrule
\multicolumn{1}{c}{Observations} & & \multicolumn{6}{c}{Learning Rate Exponent, $p$} \\ 
\multicolumn{1}{c}{used}& & $p = 1.0$ & $p = 0.9$ & $p = 0.8$ & $p = 0.7$ & $p = 0.6$ & $p = 0.5$ \\ \cmidrule{1-1} \cmidrule{3-8}
$10^2$ & & $0.0$ & $0.0$ & $0.0$ & $0.0$ & $0.0$ & $0.0$\\
$10^4$ & & $0.0$ & $0.0$ & $0.0$ & $0.0$ & $0.0$ & $0.0$\\
$10^6$ & & $0.0$ & $0.0$ & $0.0$ & $0.0$ & $0.0$ & $0.0$\\
$10^8$ & & $0.0$ & $0.0$ & $0.0$ & $0.0$& $0.327$& $0.0$ \\
$10^{10}$& &$0.0$& $0.0$ & $0.0$ & $0.996$& $0.958$& $0.583$ \\\bottomrule
\end{tabular}
\caption{Lower bounds for the probability that the SGD estimate is within an $l^2$ ball of radius $10^{-1}$ of $\beta^*$ for \cref{problem: Ideal Problem} with the learning rates used in \cref{figure:SGD Stalling}. Note, up to $10^6$ observations the lower bound on the probably remains at $0.0$. Moreover, even for $10^8$, only one of the six learning rates have non-zero lower bounds, which is reflected in \cref{figure:SGD Stalling}.}
\label{table:linear problem lower probability bounds}
\end{table}

To determine how pessimistic the bounds in \cref{corollary: SGD Convergence Rate Lower Bound} are, we can use numerical simulations. Using the same instantiation of \cref{problem: Ideal Problem} from \cref{figure:SGD Stalling}, we independently run SGD one hundred times for each learning rate used in \cref{figure:SGD Stalling}, where each run uses an independently generated set of 100 million observations; then, for each of the six hundred runs, we record error statistics. Summaries for these errors are tabulated in \cref{table:linear problem simulated runs}. Additionally, for comparison, \cref{table:linear problem simulated runs} reports the error statistics for ordinary least squares estimators under identical experimental conditions. Comparing the results of \cref{table:linear problem lower probability bounds,table:linear problem simulated runs}, it seems that \cref{corollary: SGD Convergence Rate Lower Bound} is rather conservative, but is informative about the worst-case behavior.

\begin{table}[ht]
\centering
\ra
\begin{tabular}{@{}c c rrrrrr@{}}\toprule
\multicolumn{8}{c}{SGD Convergence Summary Statistics, $d=100$, $10^8$ Observations} \\ \midrule
\multicolumn{1}{c}{Method} & & \multicolumn{6}{c}{Statistics} \\ 
  & & Mean & Median & Variance & Max & Min & Fraction \\ \cmidrule{1-1} \cmidrule{3-8}
\textit{OLS} & &  0.0287  &   0.0287 &4.7116e-6&   0.0361 &   0.0233 & 1.0 \\
             & & \\
\textit{SGD}, $p = 1.0$ & & 466.9528 & 467.8384 & 29.7957 & 477.0266 & 446.7251 & 0.0 \\
\textit{SGD}, $p = 0.9$ & & 330.2472 & 330.0679 & 13.7920 & 338.6541 & 320.6750 & 0.0 \\
\textit{SGD}, $p = 0.8$ & & 80.8682  &  80.9932 &  1.4741 &  83.0688 &  77.4359 & 0.0 \\
\textit{SGD}, $p = 0.7$ & &  0.1394  &   0.1392 &2.1745e-5&   0.1515 &   0.1282 & 0.0 \\
\textit{SGD}, $p = 0.6$ & &  0.0814  &   0.0807 &3.6933e-5&   0.1008 &   0.0640 & 0.99 \\
\textit{SGD}, $p = 0.5$ & &  0.2022  &   0.2017 &2.1454e-4&   0.2559 &   0.1625 & 0.0 \\ \bottomrule
\end{tabular}
\caption{For each of the learning rates used in \cref{figure:SGD Stalling}, a tabulation of error statistics for 100 independent SGD runs with 100 million observations each on \cref{problem: Ideal Problem}. Additionally, 100 independent OLS estimates for \cref{problem: Ideal Problem} with 100 million observations are computed. The mean, median, variance, maximum and minimum $l^2$ error for the 100 independent runs are reported. Additionally, the fraction of runs in each group which have an error below $10^{-1}$, and should be compared to the lower bounds computed in \cref{table:linear problem lower probability bounds} for $k=10^8$ observations.}
\label{table:linear problem simulated runs}
\end{table}

%

\begin{rmrk}
Another way of complementing the lower bounds in \cref{corollary: SGD Convergence Rate Lower Bound} is with an upper bound. Unfortunately. such upper bounds require particular knowledge of the underlying process or modifications of SGD which constrain it to within a fixed region containing the true parameter $\beta^*$. As such, we will not endeavor to state these results even for \cref{problem: Ideal Problem}, as they are not practical or do not generalize easily.
\end{rmrk}

Now that we have established the problem of stalling for the ideal problem \cref{problem: Ideal Problem}, we turn to stating this problem in a general case.

\subsection{A More General Problem Definition} \label{subsection: general problem}

We now consider a more general problem. Let $l:\mathbb{R}^p \times \mathbb{R}^d \to \mathbb{R}$ be a differentiable loss function, let $Z, Z_1,Z_2,\ldots \in \mathbb{R}^p$ be independent, identically distributed random variables, let the risk function be
$$ R(\beta) = \E{l(Z,\beta)}$$
Assume that $R$ is a differentiable, Lipschitz continuous gradient, strongly convex function with Lipschitz parameter $L$ and strong convexity parameter $\sigma$, and suppose that for all $\beta \in \mathbb{R}^d$,
$$ g(\beta) := \nabla R(\beta) = \E{\dot{l}(Z,\beta)}$$
In addition, assume that the variance of $\dot{l}$ has a uniformly bounded spectral norm for all $\beta \in \mathbb{R}^d$ \cite[analogous to Assumption 4.3c]{bottou2016}. Letting $\beta^* = \arg\min R(\beta)$, we can state the following problem
\begin{problem}[General Problem] \label{problem: general}
Given observations $Z_1,Z_2,\ldots$, compute $\beta^*$. 
\end{problem}
\cref{problem: general} will serve as the theoretical framework for all subsequent results in this paper.

\subsection{Mechanisms of Stalling for the General Problem} \label{subsection: mechanisms}

In order to discuss the mechanisms of stalling, we must first establish necessary conditions for convergence.

\begin{theorem} \label{theorem: Probability Convergence Bound for General Problem}
If SGD is applied to \cref{problem: general} then for any $\delta > 0$
$$ \Prb{\norm{\theta_k - \beta^*}_2 \leq \delta } \geq 1 - \frac{\norm{\theta_0 - \beta^*}_2^2\prod_{j=0}^{k}(1 - 2\alpha_k \sigma + \alpha_k^2 L^2) + C A_k}{\delta^2}$$
where $C$ is the uniform bound on the spectral norm of the variance of $\dot{l}$ and 
$$ A_k = \alpha_k^2 + \sum_{j=0}^{k-1} \alpha_j^2 \prod_{l=j+1}^k (1 - 2 \alpha_l \sigma + \alpha_l^2 L^2)$$
\end{theorem}
From \cref{theorem: Probability Convergence Bound for General Problem}, convergence readily follows using Markov's inequality:
\begin{cor} \label{corollary: Convergence for General Problem}
If SGD is applied to \cref{problem: general} and if 
$$ \limsup_{k \to \infty} A_k = 0$$
then $\theta_k$ converges to $\beta^*$ in probability.
\end{cor}

To understand the mechanisms of stalling, we must again consider \cref{figure:SGD Stalling}. From \cref{figure:SGD Stalling}, SGD makes the most progress towards $\beta^*$ using the first five percent of observations for nearly all of the learning rates. Because the observations are independently generated, it follows that the ordering of the observations has no impact on stalling. Consequently, this suggests that the only other variable quantity -- the learning rate -- must mediate stalling.

To see this more rigorously, consider the terms $A_k$ in \cref{theorem: Probability Convergence Bound for General Problem}. The quantities which dictate how quickly $A_k$ converges to zero are $\alpha_k^2$ and the decay terms 
$$ \prod_{l=j+1}^k (1 - 2 \alpha_l \sigma + \alpha_l^2 L^2) $$
From a straightforward calculation, the decay terms are minimized when $\alpha_l = \sigma/L^2$. Thus, when $\alpha_l$ are near $\sigma/L^2$, the decay terms will reduce $A_k$ quickly, whereas, as $\alpha_l$ approach zero, the decay terms no longer have an appreciable impact on causing $A_k$ to decay.

Therefore, it is ideal to set $\alpha_l = \sigma/L^2$ to optimally remove the bias terms and the effects of earlier learning rates on $A_k$, and, when $\theta_k$ approach the solution, to force the learning rate to begin to decay to $0$. Unfortunately, the well-known difficulty with such a procedure is that it will never be known when $\theta_k$ is near the optimal point, nor is it clear how ``near" should be defined as it will depend on the characteristics of the objective function at the solution. 

Our alternative approach is to allow the scheduled learning rate to decay, and then restart the learning rate so that the high-impact initial terms will result in large reductions in error. However, periodic restarts will prevent $A_k$ from converging to $0$ which is needed for convergence. Thus, the trick is to let the periods in between restarts increase sufficiently fast thereby allowing for convergence in the limit. 

We note that our approach extends the work of \cite{loshchilov2016}. \cite{loshchilov2016} were the first to numerically explore restarted SGD on deep neural networks. We extend this line of work by stating a generic restart framework which allows for random restart points, and can be applied to arbitrary stochastic gradient methods. Moreover, we prove sufficient conditions for restarted SGD to converge. We detail this approach and characterize its behavior next.

\section{Restarting} \label{section: Restart Framework}
In \cref{subsection: algorithm}, we overview the restart framework, state a restarted SGD algorithm, and numerically demonstrate its effectiveness on \cref{problem: Ideal Problem} using the same observations and conditions as used for the experiments visualized in \cref{figure:SGD Stalling}. In \cref{subsection: mathematics}, we mathematically describe the restarted SGD procedure and prove that it converges. 

\subsection{Restart Algorithm} \label{subsection: algorithm}
The restart strategy for general stochastic incremental optimizers is:
\begin{enumerate}
\item Start the procedure until a user-defined triggering event occurs. 
\item Restart the procedure at the current iterate, until another triggering event occurs.
\item Repeat this process such that the time between sequential triggering events diverges to infinity. 
\end{enumerate}
For SGD, we state a specific algorithm
\begin{program}
|input: learning rate |\lbrace \alpha_s \rbrace, |triggering events | \lbrace \tau_j \rbrace  , |initialization | \theta, |data | \lbrace Z_k \rbrace \\
k, s, j \leftarrow 1, 0, 0 \\
\WHILE true
\theta \leftarrow \theta - \alpha_s \dot{l}(Z_{k},\theta) \\
\IF \tau_j == true \THEN s, j \leftarrow 0, j+1 \FI\\
k  \leftarrow k+1
\END
\end{program}

\begin{figure}[ht]
\centering
\includegraphics[scale=0.7]{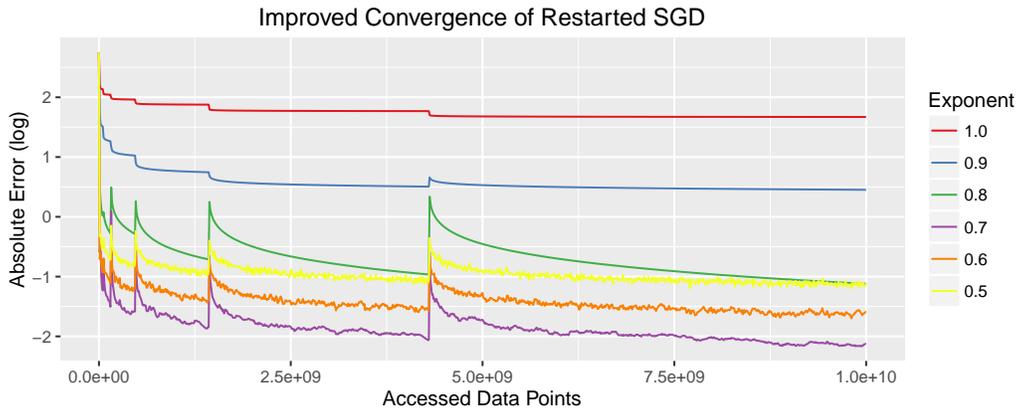}
\caption{The base-10 logarithm of the absolute error of the estimators generated by restarted SGD for the same problem and observations used in \cref{figure:SGD Stalling}. Comparing the two figures, we see that the restarted SGD drastically outperforms the standard SGD for exponents $1.0, 0.9,0.8$ and performs identically for $0.7,0.6,0.5$.}
\label{figure:SGD Restart Ideal}
\end{figure}

We apply this strategy to \cref{problem: Ideal Problem} with exactly the same problem instantiation and exactly the same data in the same order as \cref{figure:SGD Stalling}, and plot the absolute errors in \cref{figure:SGD Restart Ideal}.

Comparing \cref{figure:SGD Stalling,figure:SGD Restart Ideal}, we see that the restarted SGD variant either drastically improves on the absolute errors for the worst performing learning rates, or performs nearly identically for the best performing learning rates. We can also compare the standard and restart SGD variants under the conditions used in \cref{table:linear problem simulated runs}. We state these results in \cref{table:linear problem simulated runs restart}. 

\begin{table}[ht]
\centering
\ra
\begin{tabular}{@{}c c rrrrrr@{}}\toprule
\multicolumn{8}{c}{Restarted SGD Convergence Summary Statistics, $d=100$, $10^8$ Observations} \\ \midrule
\multicolumn{1}{c}{Method} & & \multicolumn{6}{c}{Statistics} \\ 
  & & Mean & Median & Variance & Max & Min & Fraction \\ \cmidrule{1-1} \cmidrule{3-8}
\textit{SGD}, $p = 1.0$ & & 117.9137 & 118.5475 & 31.4797 & 130.7890 & 105.7674 & 0.0 \\
\textit{SGD}, $p = 0.9$ & & 20.9212 & 21.0031 & 1.1371 & 23.5030 & 18.1436 & 0.0 \\
\textit{SGD}, $p = 0.8$ & & 0.8612  &  0.8177 &  0.0265 &  1.3330 &  0.5676 & 0.0 \\
\textit{SGD}, $p = 0.7$ & &  0.0476  &   0.0474 &1.1066e-5&   0.0563 &   0.0406 & 1.0 \\
\textit{SGD}, $p = 0.6$ & &  0.1029  &  0.1018  & 5.0506e-5& 0.1216 & 0.0851 & 0.34 \\
\textit{SGD}, $p = 0.5$ & &  0.2460  &  0.2480 & 3.4880e-4& 0.2810 & 0.1913 & 0.0 \\ \bottomrule
\end{tabular}
\caption{We repeat the simulations in \cref{table:linear problem simulated runs} using the restarted SGD variant. We see an improvement in the performance of the restarted SGD variant for learning rates with $p=1.0,0.9,0.8,0.7$ and we see a marginal decline in performance for $p=0.6,0.5$.}
\label{table:linear problem simulated runs restart}
\end{table}


Comparing \cref{table:linear problem simulated runs,table:linear problem simulated runs restart}, we see a drastic improvement for $p=1.0,0.9,0.8,0.7$ by a factor of $2.9$ to $93.9$, but a marginal decline in performance for $p=0.6$ and $p=0.5$ by a factor of $1.1$ to $1.4$. Therefore, on the ideal problem, we have a great deal of evidence which suggests that the restart strategy will improve the overall behavior of SGD. We now give the mathematical details for this method.

\subsection{Mathematical Details} \label{subsection: mathematics}

Fix the learning rate $\lbrace \alpha_k : k+1 \in \mathbb{N} \rbrace$. Moreover, re-index $Z_1,Z_2,\ldots$ with two indices 
$$Z_{0,1}, Z_{0,2}, \ldots, Z_{1,1}, Z_{1,2},\ldots,Z_{j,1},Z_{j,2},\ldots $$
Now, let $\lbrace \theta_{j,k} : j+1, k+1 \in \mathbb{N} \rbrace$ be random variables where $\theta_{0,0}$ is arbitrary and for $k +1 \in \mathbb{N}$ are defined by
$$ \theta_{j,k+1} = \theta_{j,k} - \alpha_k \dot{l}(Z_{j,k+1},\theta_{j,k})$$
In order to specify $\theta_{j,0}$, we will need to define the triggering events, which we will do using stopping times. Let $\mathcal{F}_{j,k} = \sigma(\theta_{j,0},\ldots,\theta_{j,k})$, $\tau_0 = 0$, and let $\tau_{j+1}$ be stopping times with respect to $\lbrace \mathcal{F}_{j,k}:k+1\in \mathbb{N} \rbrace$ such that
\begin{enumerate}
\item $\tau_{j} < \tau_{j+1}$ and $\tau_{j} \to \infty$ almost surely.
\item $\tau_{j+1}$ is independent of $\norm{\theta_{j,k} - \beta^*}_2$ for all $k \in \mathbb{N}$. 
\end{enumerate}  
This last requirement ensures that $\tau_{j}$ cannot be based on any unrealistic, oracle knowledge of the true parameter $\beta^*$. Now, let
$$ \theta_{j+1,0} = \theta_{j,\tau_{j+1}}$$
Moreover, for any $n + 1 \in \mathbb{N}$ define
$$ S_j = \sum_{l=0}^j \tau_l \quad\text{and}\quad J(n) = \sup \left\lbrace j: S_j \leq n \right\rbrace$$
With this notation, we define restarted SGD to be the sequence of estimates 
\begin{equation} \label{equation: restarted SGD}
\lbrace \theta_n = \theta_{J(n),n - S_{J(n)}} : n +1 \in \mathbb{N} \rbrace
\end{equation}

\begin{theorem} \label{theorem: Restart SGD Convergence}
If restarted SGD is applied to \cref{problem: general}, and if the limit supremum of the sequence
$$ A_k := \alpha_k^2 + \sum_{j=0}^{k-1} \alpha_j^2 \prod_{l=j+1}^k (1 - 2 \alpha_l \sigma + \alpha_l^2 L^2)$$
converges to $0$, then $\theta_{S_j}$ converges to $\beta^*$ in probability as $j \to \infty$. 
\end{theorem}

\begin{remark}
It is most likely beneficial to cycle through a variety of learning rates at each restart in order to get the maximal benefit of each learning rate on the problem. This extension is straightforward in terms of analysis and in implementation. However, we will not consider it further as our goal is to demonstrate that the simple restart strategy offers a dramatic improvement in performance of SGD and its variants.
\end{remark}

\section{Numerical Experiments} \label{section: Experiments}
In \cref{subsection: data task model}, we describe the data set, the preprocessing, the learning task and the model used to satisfy the task. In \cref{subsection: experimental set up}, we detail the computing environment, the optimization methods and their parameters, and the output metrics for comparing the standard methods against their restarted analogues. In \cref{subsection: results and discussion}, we report and discuss the results. 

\subsection{Data Set, Task and Model} \label{subsection: data task model}

In the experiments, we use a data set available on the UCI Repository which was generated by a Fermi Lab experiment used to test techniques for differentiating between electron neutrinos, considered the signal, and muon neutrinos, considered the background \citep{roe2005}. The data set contains 130,064 examples where the first 36,499 examples correspond to electron neutrinos, and the remaining 93,565 correspond to muon neutrinos, and each example has a $p = 50$ dimensional feature vector. The task is set to discern between the electron and muon neutrinos using the feature vector. 

The data set was preprocessed by prepending a $1$ to the example if it corresponded to a signal, and a $0$ otherwise. The data set was randomly shuffled and 91,044 examples (approximately seventy percent) of the data was used as the training set, and the remaining 39,020 examples were left as the testing set. 

The former data set was used to train a two-layer feed forward neural network with the following architecture:
\begin{enumerate}
\item The observation layer contained one neuron with a logistic activation function with five inputs and one output.
\item The hidden layer contained five neurons, each with a logistic activation function. The output of these five neurons fed into the observation layer neuron. Each of the five neurons was arbitrarily assigned ten of the feature vectors without overlap.
\end{enumerate}
The resulting model had $d = 61$ dimensional parameter vector to be learned.

\subsection{Experimental Set Up} \label{subsection: experimental set up}

The experiments listed below were run on a machine with an Intel i5 Processor (3.33 GHz) with nearly 4 GB of memory. 

Seven methods were used to train the model using the training data with exactly 30 epochs and were all initialized at exactly the same random value: SGD, restarted SGD, AdaGrad, restarted AdaGrad, kSGD, restarted kSGD, and BFGS. As the goal of the experiment is to compare standard methods against their restarted analogies, there was little effort to optimize the hyperparameters for the methods. The learning rate for SGD and restarted SGD were arbitrarily set to $$\alpha_l = l^{-0.7}$$
For AdaGrad and restarted AdaGrad, the multiplicative factor was set to $\eta = 0.001$. For restarted AdaGrad, the restart reset the adaptive learning rate to the vector of ones. For kSGD and restarted kSGD, the hyperparameter was set to $\gamma^2 = 0.01$. For restarted kSGD, the restart reset the covariance estimate to the identity. For BFGS, the maximal line search length was $\alpha_0 = 1$, the line search reduction factor was $\rho = 0.5$, and the Armijo condition parameter was set to $c = 0.0001$.

The triggering events were set to deterministic values. In particular the first triggering event occurred at iteration $100$. All future triggering iteration occurred a factor of $1.56$ times the previous triggering iteration. The factor of $1.56$ was selected to ensure a large number of restarts occurred (each method restarted 22 times) and that the method was just shy of another restart in order to see the impact of the restarts. This factor was selected before any testing was done.

For each of the stochastic methods, the parameter was recorded every 5,000 iterations and at the last iteration. For BFGS, the parameter was recorded at the end of each iteration. For each recorded parameter, the training and testing error were computed. For the last recorded parameter, the total gradient was computed.

\subsection{Results and Discussion} \label{subsection: results and discussion}
\begin{figure}
\centering
\includegraphics[scale=0.7]{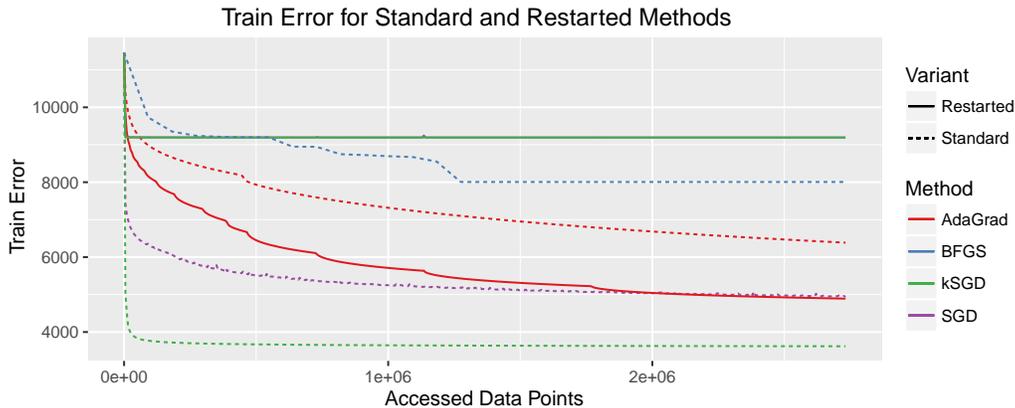}
\caption{Training error for the neural network model on the neutrino data which was learned using the standard SGD, AdaGrad, kSGD and BFGS methods, and restarted SGD, AdaGrad and kSGD methods.}
\label{figure: train_error}
\end{figure}

\begin{figure}
\centering
\includegraphics[scale=0.7]{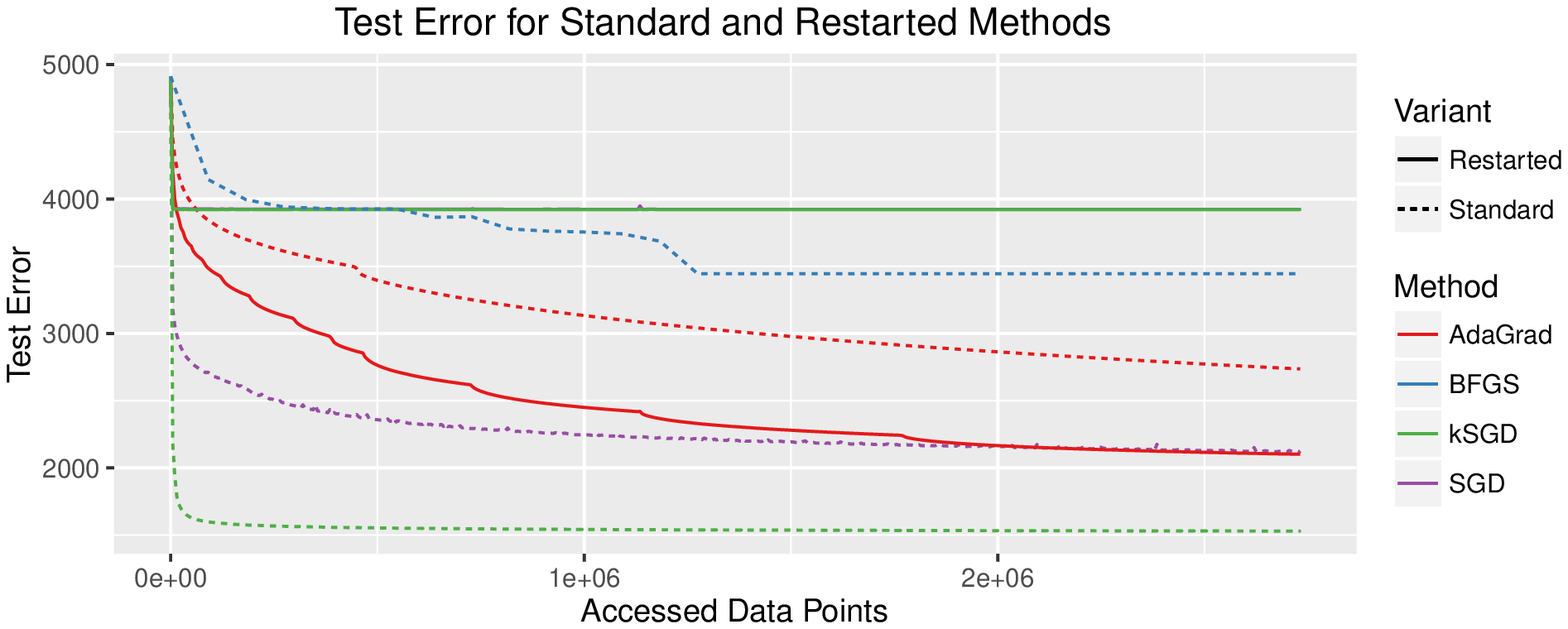}
\caption{Testing error for the neural network model on the neutrino data which was learned using the standard SGD, AdaGrad, kSGD and BFGS methods, and restarted SGD, AdaGrad and kSGD methods.}
\label{figure: test_error}
\end{figure}

In \cref{figure: train_error,figure: test_error}, the training and testing error for the different methods are plotted. From these figures, it seems that the restart methods for SGD and kSGD do much worse than their standard counterparts, while the restart method for AdaGrad outperforms its standard counterpart. However, the non-linear nature of the neural network and the local search behavior of these learning methods is confounding the results. As evidenced by the training error and testing error of BFGS, many of these methods, owing to their random sampling nature, may end up in rather different local minima which have different training and testing errors. Therefore, the correct quantity to consider is the norm of the total gradient of the model on the training error for the final measured parameter. This quantity describes if a stationary point has actually been found by the different learning methods. These values are tabulated in \cref{table: train_gradient_errors}. Thus, in light of \cref{table: train_gradient_errors}, the restarted SGD and kSGD variants perform quite well: indeed, both seem to converge very quickly to their local stationary points in comparison to their standard variants. Similarly, the restarted AdaGrad also has a better total gradient norm in comparison to the standard AdaGrad method.

\begin{table}
\centering
\ra
\begin{tabular}{@{}l c r r r r @{}}\toprule
\multicolumn{6}{c}{Comparison of Total Gradients for Neutrino Problem} \\ \midrule
\multicolumn{1}{c}{Variants} & & \multicolumn{1}{c}{SGD} & \multicolumn{1}{c}{AdaGrad} & \multicolumn{1}{c}{kSGD} & \multicolumn{1}{c}{BFGS} \\ \cmidrule{1-1} \cmidrule{3-6}
Standard & & 0.08176 & 0.01909 & 0.05436 & 0.02065 \\
Restart & &  0.00015 & 0.01018 & 8.176e-6 & \multicolumn{1}{c}{---} \\\bottomrule
\end{tabular}
\caption{The gradient of the training objective function for the final parameter of each learning method. Here, we see that the restart methods are much closer to the respective stationary points in comparison to the standard methods.}
\label{table: train_gradient_errors}
\end{table}


\section{Conclusions} \label{section: Conclusion}
In this work, we mathematically and numerically demonstrated that stalling is a severely limiting and generic property of Stochastic Gradient Descent and its variants. Then, we introduced a simple restart strategy for hedging against the impact of stalling for these stochastic gradient methods. We proved that the restarted variant of SGD will still converge, and numerically demonstrated its effectiveness on an ideal risk minimization problem. Finally, we showed that the restart strategy can be applied to other stochastic incremental optimization methods by using the examples of AdaGrad and kSGD on a neural network model. We experimentally verified that the restart strategy when applied to SGD, AdaGrad and kSGD improves convergence to the local stationary point. 

One future direction of this work is to understand how the local problem geometry, variability of the stochastic gradients, and restart strategy may actually force the parameters to explore other regions of the objective function's manifold and provide guarantees of avoiding stationary points which are not minimizers. Another future direction for this work is to use the restart strategy to provide generic, high-probability stopping criteria for a variety of stochastic incremental techniques including SGD and AdaGrad.


\acks{The author is supported by the Department of Energy Office of Science through contract DE-AC02-06CH11357, and the National Science Foundation through awards FP061151-01-PR and CNS-1545046. The author would like to thank Mihai Anitescu for his advice, conversations and feedback on this work.}

\newpage

\appendix
\section{Proofs}
\begin{proof}{\bf \cref{theorem: SGD Convergence for Ideal Problem}.} The first part follows by a direct computation using the properties of \cref{problem: Ideal Problem}:
\begin{align*}
d\left(\E{R(\theta_{k+1})} - R(\beta^*)\right) &= \E{\norm{\theta_{k+1} - \beta^*}_2^2} \\
 &= \E{\norm{\theta_k - \beta^* - \alpha_k X_{k+1}X_{k+1}'(\theta_k - \beta^*) + \alpha_k X_{k+1} \epsilon_{k+1} }_2^2} \\
 &= \E{\norm{\theta_k - \beta^*}_2^2} + \alpha_k^2 \V{\epsilon}  - (2\alpha_k - \alpha_k^2) \times\\
 &\quad \E{\cond{\tr{(\theta_k - \beta^*)(\theta_k - \beta^*)'X_{k+1}X_{k+1}'}}{\sigma(\theta_k)}} \\
 &= \E{\norm{\theta_k - \beta^*}_2^2} + \alpha_k^2 \V{\epsilon} - (2\alpha_k - \alpha_k^2) \times \\
 &\quad \E{\tr{(\theta_k - \beta^*)(\theta_k - \beta^*)'\E{X_{k+1}X_{k+1}'}}} \\
 &= \E{\norm{\theta_k - \beta^*}_2^2}\left(1 + \frac{\alpha_k^2 - 2\alpha_k}{d} \right)^2 + \V{\epsilon} \alpha_k^2
\end{align*} 
Using this equation recursively, we have the first part of the result. Using the first part of the result and Markov's inequality, the condition for convergence of $\theta_k$ to $\beta^*$ in probability is readily proved.
\end{proof}
\begin{proof}{\bf \cref{corollary: SGD Convergence Rate Lower Bound}.} From Markov's Inequality and \cref{theorem: SGD Convergence for Ideal Problem},
\begin{align*}
\Prb{\norm{\theta_k - \beta^*}_2 > \delta} &= \Prb{\norm{\theta_k - \beta^*}_2^2 > \delta^2} \\
											&= \Prb{R(\theta_k) - R(\beta^*) > \frac{\delta^2}{d}} \\
											&\leq d\frac{\left[R(\theta_0) - R(\beta^*) \right]\prod_{j=0}^k \left(1 + \frac{\alpha_j^2 - 2\alpha_j}{d} \right) + \frac{\V{\epsilon}}{d}A_k}{\delta^2} \\
											&= \frac{\norm{\theta_0 - \beta^*}_2^2 \prod_{j=0}^k \left(1 + \frac{\alpha_j^2 - 2\alpha_j}{d} \right) + \V{\epsilon}A_k}{\delta^2}
\end{align*}
Computing the probability of the complement gives the result.
\end{proof}
\begin{proof}{\bf \cref{theorem: Probability Convergence Bound for General Problem}.}
Let $\mathcal{F}_k = \sigma(\theta_1,\ldots,\theta_k)$ be a filtration. Then
\begin{align*}
\cond{\norm{\theta_{k+1} - \beta^*}_2^2}{\mathcal{F}_k}
 &= \norm{\theta_{k} - \beta^*}_2^2 - 2\alpha_{k}(\theta_k - \beta^*)'g(\theta_k) + \alpha_{k}^2 \cond{ \norm{\dot{l}(Z_{k+1},\theta_k)}_2^2}{\mathcal{F}_k}
\end{align*}
By optimality, $g(\beta^*) = 0$. Therefore, by optimality and strong convexity
\begin{align*}
(\theta_k - \beta^*)'g(\theta_k) &= (\theta_k - \beta^*)'(g(\theta_k) - g(\beta^*)) \\
								&\geq \sigma \norm{\theta_k - \beta^*}_2^2 
\end{align*}
Also, using $g(\beta^*) = 0$, Lipschitz continuity of $g$, and the uniform variance bound
\begin{align*}
\cond{\norm{\dot{l}(Z_{k+1},\theta_k)}_2^2}{\mathcal{F}_k} &= \cond{\norm{\dot{l}(Z_{k+1},\theta_k) - g(\theta_k)}_2^2}{\mathcal{F}_k} + \norm{g(\theta_k) - g(\beta^*)}_2^2 \\
 &\leq C + L^2 \norm{\theta_k - \beta^*}_2^2
\end{align*}
Putting these components together
$$\cond{\norm{\theta_{k+1} - \beta^*}_2^2}{\mathcal{F}_k} \leq (1 - 2\alpha_k \sigma + \alpha^2 L^2 ) \norm{\theta_k - \beta^*} + \alpha_k^2 C$$
Taking expectations of both sides, and using the result recursively gives
$$ \E{\norm{\theta_{k+1} - \beta^*}_2^2} \leq \norm{\theta_0 - \beta^*}_2^2\prod_{j=0}^{k}(1 - 2\alpha_k \sigma + \alpha_k^2 L^2) + C A_k$$
Now, using this with Markov's Inequality, as done in the proof of \cref{corollary: SGD Convergence Rate Lower Bound}, gives the result.
\end{proof}
\begin{proof}{\bf \cref{theorem: Restart SGD Convergence}.} We will compute $\E{\norm{\theta_{S_j} - \beta^*}_2^2}$, and show that its limit converges to $0$. Let
$$ \gamma_j := \prod_{l=0}^{j} 1 - 2 \alpha_l \sigma + \alpha_l^2 L^2 $$
and consider
\begin{align*}
&\cond{\norm{\theta_{S_{j+1}} - \beta^*}_2^2 }{\sigma(\theta_{S_{j}},\ldots,\theta_{S_{j+1}-1},S_{j},S_{j+1})} \\
&= \cond{\norm{\theta_{S_{j+1}-1} - \beta^* - \alpha_{\tau_{j+1}-1} \dot{l}(Z_{j,\tau_{j+1}},\theta_{S_{j+1}-1})}_2^2}{\sigma(\theta_{S_{j}},\ldots,\theta_{S_{j+1}-1},S_{j},S_{j+1})}\\
&\leq \norm{\theta_{S_{j+1}-1} - \beta^*}_2^2(1 - 2\alpha_{\tau_{j+1}-1}\sigma + 2 \alpha_{\tau_{j+1}-1}^2 L^2) + C \alpha_{\tau_{j+1}-1}^2
\end{align*}
Repeating this computation, we have
$$\cond{\norm{\theta_{S_{j+1}} - \beta^*}_2^2}{\sigma(S_{j},S_{j+1})} \leq \cond{\norm{\theta_{S_{j} }- \beta^*}_2^2}{\sigma(S_j)} \gamma_{\tau_{j+1}-1} + C A_{\tau_{j+1}-1}$$
Iterating over this result,
\begin{align*}
&\cond{\norm{\theta_{S_{j+1}} - \beta^*}_2^2}{\sigma(S_0,\ldots,S_{j+1})} \\
& \leq 
\left[ \cdots \left[ \left[\norm{\theta_{0} - \beta^*}_2^2 \gamma_{\tau_1 - 1} + C A_{\tau_1 - 1}  \right] \gamma_{\tau_2 - 1} + C A_{\tau_2-1}\right] \cdots \right] \gamma_{\tau_{j+1}-1} + C A_{\tau_{j+1} - 1}
\end{align*}
By assumption on $A_k$, $\exists J_1 \in \mathbb{Z}_{\geq 0}$ such that
\begin{enumerate}
\item $\sup_{j \geq J_1} \gamma_{j} =: \tilde{\gamma} < 1$
\item $J_1$ is the minimal such integer
\end{enumerate}
Let $\bar{\gamma} = \sup_{j} \gamma_j$ and $\bar{A} = \sup_{j} A_j$. Recalling, by construction of $\tau_j$, $j \leq \tau_j$, then for any $j > J_1$
\begin{align*}
\cond{\norm{\theta_{S_{j+1}} - \beta^*}_2^2}{\sigma(S_0,\ldots,S_{j+1})} &\leq \left[ \norm{\theta_{0} - \beta^*}_2^2 \bar{\gamma}^{J_1} + \bar{A} \sum_{l=0}^{J_1} \bar{\gamma}^{l}  + \bar{A} \frac{1}{1 - \tilde{\gamma}} \right] \gamma_{\tau_{j+1} - 1} \\ 
&+ C A_{\tau_{j+1} - 1}
\end{align*}
Hence,
\begin{align*}
 \E{\norm{\theta_{S_{j+1}} - \beta^*}_2^2} &\leq \left[ \norm{\theta_{0} - \beta^*}_2^2 \bar{\gamma}^{J_1} + \bar{A} \sum_{l=0}^{J_1} \bar{\gamma}^{l}  + \bar{A} \frac{1}{1 - \tilde{\gamma}} \right] \E{\gamma_{\tau_{j+1} - 1}}\\ 
&+ C \E{A_{\tau_{j+1} - 1} }
\end{align*}
By the dominated convergence theorem,
$$ \limsup_{j \to \infty} \E{\norm{\theta_{S_{j}} - \beta^*}_2^2} = 0$$
By Markov's inequality, the result follows.
\end{proof}
\vskip 0.2in
\bibliography{bibliography}

\end{document}